# Do You Think It's Biased?

# How To Ask For The Perception Of Media Bias

Timo Spinde
*University of Konstanz*
Konstanz, Germany
Timo.Spinde@uni-konstanz.de

Christina Kreuter
*University of Konstanz*
Konstanz, Germany
Christina.kreuter@uni-konstanz.de

Wolfgang Gaissmaier
*University of Konstanz*
Konstanz, Germany
Gaissmaier@uni-konstanz.de

Felix Hamborg
*University of Konstanz*
Konstanz, Germany
Felix.hamborg@uni-konstanz.de

Bela Gipp
*University of Wuppertal*
Wuppertal, Germany
gipp@uni-wuppertal.de

Helge Giese
*University of Konstanz*
Konstanz, Germany
Helge.Giese@uni-konstanz.de

***Abstract*— Media coverage possesses a substantial effect on the public perception of events. The way media frames events can significantly alter the beliefs and perceptions of our society. Nevertheless, nearly all media outlets are known to report news in a biased way. While such bias can be introduced by altering the word choice or omitting information, the perception of bias also varies largely depending on a reader's personal background. Therefore, media bias is a very complex construct to identify and analyze. Even though media bias has been the subject of many studies, previous assessment strategies are oversimplified, lack overlap and empirical evaluation. Thus, this study aims to develop a scale that can be used as a reliable standard to evaluate article bias. To name an example: Intending to measure bias in a news article, should we ask, "How biased is the article?" or should we instead ask, "How did the article treat the American president?". We conducted a literature search to find 824 relevant questions about text perception in previous research on the topic. In a multi-iterative process, we summarized and condensed these questions semantically to conclude a complete and representative set of possible question types about bias. The final set consisted of 25 questions with varying answering formats, 17 questions using semantic differentials, and six ratings of feelings. We tested each of the questions on 190 articles with overall 663 participants to identify how well the questions measure an article's perceived bias. Our results show that 21 final items are suitable and reliable for measuring the perception of media bias. We publish the final set of questions on http://bias-question-tree.gipplab.org/.**

## I. Introduction

News media play a fundamental role in the democratic process. Many people consider news "articles a reliable source of information about current events, even though it is also broadly believed and academically confirmed that news outlets are biased" [1]. Given the trust readers put into news articles "and the significant influence of media outlets on society and public opinion, media bias may potentially lead to the adoption of biased views by readers" [2]. The news, therefore, play an essential part in forming public opinion on political and other current issues [3]. Simultaneously, unrestricted access to unbiased information about any topic is crucial to develop a balanced viewpoint on different events [1]. The severity of biased news coverage is "amplified further by the fact that regular news consumers are typically not fully aware of its degree and scope" [2].

Even though media bias and its perception are prevalent and relevant issues in society and research, its reliable measurement poses a challenge. Recent computer scientific research aiming to build automated media bias detection systems reported that building a high-quality bias data set is difficult because readers struggle to agree on biased text documents [1], [4]. Many individual factors affect the perception of bias, such as topic knowledge, political ideology, or simply age and education [1]. Phenomena like the Hostile Media Effect (HME, describing the tendency to perceive media coverage of an issue as biased against one's views [5]) might also play a role, making it hard to objectively determine whether and how an article or clip is biased.

Throughout the different studies on media bias perception and identification, various definitions and methods were used to measure media bias. Still, there is a major lack of agreement about how study participants or readers react towards bias depending on how they were asked. Most existing studies focus only on specific aspects, for example, the already mentioned HME [6]–[8]. Some studies asked questions related to particular articles [9],

This work was supported by the Hanns-Seidel-Foundation and the Federal Ministry of Education and Research of Germany.

while others chose a more general approach [10]. Some ask about bias directly (e.g., "Regarding the web page that you viewed, would you say the portrayal of the presidential candidates was strictly neutral or biased in favor of one side or the other? "[11]), and some indirectly [2], [12], [13]. Some researchers tried experiments [14], while others use surveys [10].

While there is some overlap in questions across multiple studies (for example, questions similar to "Would you say that the content in this article was strictly neutral, or was it biased in favor of one side or the other?" [15] were used in different studies [11], [16], [17]), there is a large variety in methods and definitions used in prior research that limits studies' comparability on media bias perception. Furthermore, a standard of assessing media bias of articles as a general construct is essential to train automated classifiers or build data sets: Without a clear measurement of the construct, no classifier in the related areas can reach its full potential. Our project, therefore, aims to develop questions that can be used as a reliable standard to perform new analyses or reevaluate past studies, independent of the research area.

Our primary goal and contribution is to develop a reliable scale to evaluate articles in terms of media bias. We, therefore, conducted a literature review to find 824 relevant questions about text perception in previous research on the topic, which we summarized and condensed in a multi-iterative process to a final set of 48 questions. We reduced the number of questions even more and uncovered communalities between questions empirically using exploratory factor analysis (EFA), a data reduction approach. We assess the perception of bias among various articles with a known bias rating, given different questions. The scale aims to improve the data collection on media bias. This paper describes the question testing process and summarizes and transparently visualizes the question set.

We organize the rest of the paper as follows: First, in Section 2, we present a literature review on existing studies about media bias. In Section 3 we describe our methodology, followed by our results in Section 4. Finally, we give an outlook on future work and a summary of the current project in the sections 5 and 6. We want to mention that we use the words "question" and "item" interchangeably during our work.

## II. Related Work

### A. Public bias ratings

Different platforms try to address media bias in news outlets. For instance, the news aggregator Allsides publishes bias ratings for various news outlets[1]. The bias rating by Allsides represents subjective judgments made by their readers. They are organized in five classes [18]:

*Left - Lean Left - Center - Lean Right - Right.*

Allsides combines different methods to create their ratings. They indicate which outlets and articles have been evaluated with which methods on each source page [18]. Altogether, they use the following methods[2]:

1. *Blind Bias Survey*. Allsides gathers readers "from all parts of the political bias spectrum to read and rate articles and headlines blindly — without telling them the source of the content. (…) To assure that the survey audience reflects the social and political diversity of the US, they then normalize the data" [18].
2. *Editorial Review*. To some extent, the Allsides editorial staff reviewed "the works of any source. The reviews always include a diversity of individuals covering the full range of political bias from left to right" [18].
3. *Third-Party Analysis*. The third-party "analysis may include academic research, surveys, or analysis from third parties that have a published and transparent system for evaluating the bias of multiple sources" [18].
4. *Independent Review*. An AllSides "editor, or multiple editors, reviewed content from this source and came to a general conclusion on its bias; they also investigated what the media and other sources, both partisan and nonpartisan, reported about the political leanings of this source. This method is frequently used for initial bias ratings before more robust methods can be applied, or ratings for which the bias of an outlet is relatively easy to discern" [18].
5. *Community Feedback*. "For every article posted on Allsides, a user can indicate whether or not he agrees with the ratings. While the ratings are not determined by community votes. they are used to check the performance of the current ratings." [18]

### B. Literature about bias

In various research areas, text perception and particularly bias detection have been investigated. For example, the influence of biased reporting towards citizens' use of traditional, citizen, and social media has been researched [19]. Other projects focused on hostile media perceptions [20], the influence of user-related variables on the perceptions of bias [8], [9], [15], [21]–[24], or the perception of bias in particular topics [10]. Topic-dependent text perception [14], user comments [11], [25], and visual features [26] were also main interests in the existing research.

Apart from political or communication studies and psychology, there has especially been an increasing number of computer science publications about the automated detection of media bias or the related concepts of framing and sentiment analysis [1], [4], [12], [13], [27]–[33].

[1] See https://www.Allsides.com/unbiased-balanced-news, accessed on 2021-01-08.

[2] More about the methods can be found on https://www.allsides.com/media-bias/media-bias-rating-methods

Independent of the research area, all the research mentioned above questioned either students, experts, or crowdsource workers about their perception of bias on a word, sentence, article, or image level. However, almost none reported a detailed process description on how they created the respective evaluation surveys or chose the questions that were handed to the participants. Also, especially in the computer science studies, except for the study by Spinde et al. [1], none asked for the participant's personal background. Still, as shown in some of the work from psychology and communication science, the personal background seems to be crucial information needed to understand how to interpret and use the collected feedback annotations. The data sets used in the various computer scientific approaches and projects did not reflect media bias's complexity. Instead, they primarily focused on technical approaches. We believe that bias can only be uncovered in an interdisciplinary approach and that data quality and comparability play a crucial role in training any classifier. Therefore, even more, a common and reliably evaluated question set is necessary.

## III. Methodology

### A. Literature search

To systematically find items relevant to media bias perception, we conducted an extensive search on PsychInfo and Google Scholar. Mainly, the search term "Perception of Media Bias" was used to identify relevant studies on both literature platforms. We excluded articles in languages other than English and German. We manually screened headlines and keywords for their connection to media, media bias, and media perception. If in doubt, we included articles to avoid missing relevant studies. From an original set of 405 potentially relevant papers, after extensive reading and abstract checking, we excluded all but 107 studies, for which we tried to obtain full texts. We excluded 29 more because the full-text reading showed a non-sufficient connection to the perception of media bias. We excluded another 17 studies because they did not use any items on the perception of media or media bias. Overall, we included 74 studies in our collection to create our questionnaire on media bias perception[3].

### B. Item collection and selection

Our paper collection led to a list of media bias and related variables that included the item's source, the response format if mentioned, and other important information. If available, we copied the original items from the supplementary material provided by the authors. If no supplemental materials were available, we extracted items from the articles' method and results sections. When the original wording of the item was named, the original wording was added to the list. If not, we used the provided description to reconstruct the wording as good as possible. This process resulted in a list of 824 items, which we then continued to reduce and filter in a process of three iterations. We illustrate the process in Fig.1. It is based on four main criteria, which we will summarize afterward:

1. The items relate to media bias.
2. The items cover different aspects of media bias.
3. The items measure media bias on an article level.
4. The items are usable for visual analog scales (VAS[4]).

At first, in the categorization iteration, we organized the questions into general categories (e.g., Political Background, Demographics, Perception of Media Bias, Influence of Media Bias). We only included items categorized into "Perception of Media Bias" and "Influence of Media Bias" for creating a list of potential items (419). The other categories were revisited later to find relevant background information items, for example, on demographics or political background.

To further reduce the number of items for assessment, we grouped items into the following bias measurement categories: Cause, Existence, Direction, Strength, and Influence. We then grouped semantically and topically similar items to find a construct that fitted as many items as possible without losing any relevant aspects. To name an example, one of the resulting constructs was: "Would you say that the "person/content/outlet was strictly neutral, biased against, or in favor of "side"? Overall, 42 constructs and 99 general items without constructs were left after this process. Since 141 items were still too many, we grouped the edited items by their content and chose items to cover every aspect of each content in a final iteration. If possible, we selected a construct. If a construct did not cover an aspect, we used one of the general remaining items. As a result of this process, we had to exclude some items for the following reasons:

1. We decided on a visual analog scale as the response format for the questionnaire. Most questions could be adapted to fit this format, but we removed questions where this was not possible.
2. Since the questionnaire is supposed to identify bias in an article, some questions were too unspecific or unfit for this questionnaire, for example, questions about media outlets.
3. Many studies did not include the original wording of their items, and in a few cases, it was not possible to create an adequate item out of the description given in the text.
4. Some items were too specific to the issue of their original study and were unfit to be included in a general questionnaire.

---

[3] They are included in the file upload at https://zenodo.org/record/4651186#.YGR5vD9CRxs and in the tree visualization on http://bias-question-tree.gipplab.org/, which we describe both in the remainder of this paper.

[4] A Visual Analogue Scale (VAS) is a measurement instrument that tries to measure a characteristic or attitude that is believed to range across a continuum of values and cannot easily be directly measured [34] .

5. Various studies used semantic differentials to ask for their respondents' impressions of the articles. In the questionnaire, we only included semantic differentials that at least two different authors used. We applied the same procedure to questions on feelings. We excluded some items because they were only used once.

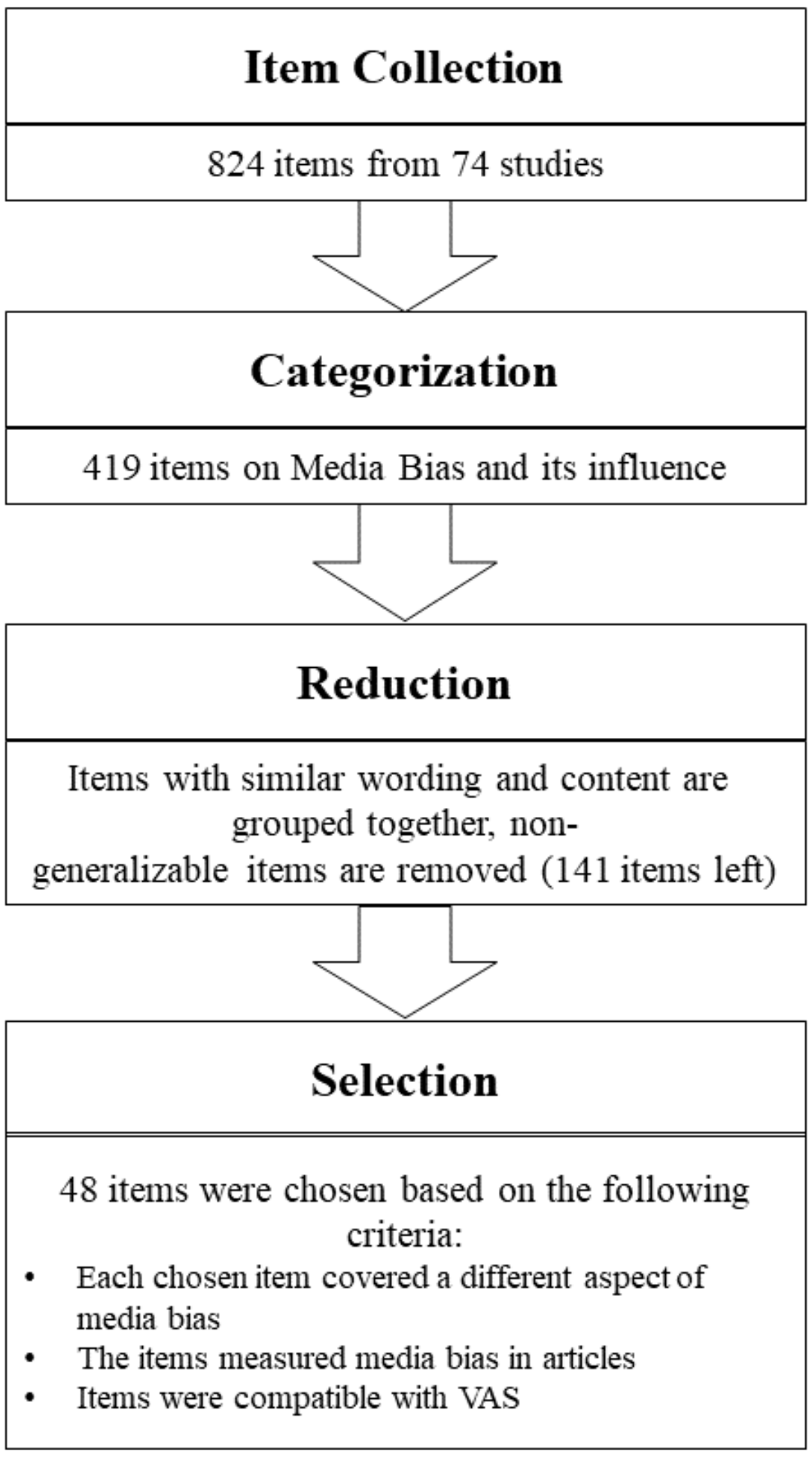

Fig. 1. Item reduction process in four main phases.

After this selection, exclusion, and merging process, the final questionnaire consisted of 25 items with varying answering formats, 17 semantic differentials, and six ratings of feelings. To cover third-person perception, we included three items twice, once asking about the article's impact on the participant directly and once asking about the impact on others. For the question about others, we used the term "another person" to keep the questionnaire as general as possible, as performed likewise in other research [17], [22], [35], [36]. Five items remained with the placeholder that was replaced with article-specific information.

We publish the complete set of final questions and original questions, and all other process information at https://zenodo.org/record/4651186#.YGR5vD9CRxs. We also illustrate which questions were merged and excluded in which way in an interactive tree visualization on http://bias-question-tree.gipplab.org/.

## C. Design

We used the survey platform UniPark [5] for data collection and recruited participants via the recruiting platform Prolific [6]. The study ran on Oct. 20, 2020. Participants were welcomed to the study and given general information on the study's purpose and the data handling. After agreeing to participate, each participant read one of the 190 articles, which was randomly selected. We then asked each participant to rate all 48 items on five pages, separated based on differing anchors, on VAS. All VAS in the study ranged from –10 to 10 and recorded only integer numbers. The order in which the pages and the items on each page were presented was randomized. In addition, an item that asked whether participants read the article was mixed in as an attention check.

After rating the article, the participants were asked to answer general media bias questions and give demographic and background information. At the end of the study, we asked them whether their data could be used for scientific purposes, and a chance to comment on the study was given.

## D. Survey participants

We recruited a sample of 940 American participants, of which 827 participated in the study. We had to exclude 91 because of missing data. We excluded another 18 participants who indicated that their data could not be trusted and further 55 participants that indicated that they had not read the article (i.e., not the highest quarter of the rating scale). The final sample consisted of 663 participants (53.5 % women, 44.8 % men, 1.7 % other). The mean age was 33.86 (SD=13.35), ranging from 18 to 80 years. The highest level of education of participants ranged from some high school education (1.1 %), high school graduate (10.9 %), vocational or technical school (1.2 %), some college education (24.3 %), an associate degree (8.6 %), and a bachelor's degree (35.7 %), to graduate work (18.3 %). On average, participants reported spending 2.95 hours per day viewing or reading the news (SD=3.87). All participants volunteered for the study and gave informed consent. We estimated the duration of this study at 12 minutes. After completing the study, participants received £1.50 as payment. Participants described themselves as tending to be politically interested (M = 2.76, SD = 5.75) and modestly politically involved (M = –0.45, SD = 5.46.). The average self-reported political orientation leaned towards liberalism (M=–2.89, SD=5.43; –10=very liberal, 10=very conservative), and there was no clear agreement to the general existence of media bias (M = –0.51, SD = 4.24).

## E. Article Selection

Regarding the articles each participant read, we followed the article selection process described in [1] to create a sample that balances the number and extremity of both politically left and right articles included. We chose 190 articles from different topics, media outlets, authors, writing styles, and most importantly, articles that range from unbiased to very biased and politically lean towards different sides.

[5] https://www.unipark.com/, accessed on 2021-01-08.

[6] https://www.prolific.co/, accessed on 2021-01-08.

To select such a sample, we obtained the articles for this study from the platform Allsides. Out of the various topics that Allsides covers, we chose ten different topics to cover a broad spectrum based on two parameters: Current issues (e.g., Coronavirus, Elections) versus general topics (e.g., Economy, Racism), and controversial (e.g., Gun Control, Abortion, Immigration) versus less controversial topics (e.g., Arts and Entertainment, Disasters, World News). From each of the ten topics, we chose 17 articles, six articles biased to the left (three left, three lean left), five articles rated center, and six articles biased towards the right (three lean right, three right). Therefore, we overall collected 170 articles for this study from Allsides. To extend our data set with rather extreme content, we added another 20 articles, ten extremely left, ten extremely right (two for each topic), directly from alternative news outlets. The extended Allsides ratings of political ideology thus ranged from very liberal (1) to very conservative (7) (M = 4, SD = 1.59; ratings adjusted to include the ten extreme articles of either side). The articles varied among outlets and were published between Oct. 1, 2019, and Oct. 31, 2020, and are all under 1500 words long. To avoid confounding variables, we showed only plain texts. We present a complete list of articles, their ratings, further information, and their issue statements on https://zenodo.org/record/4651186#.YGR5vD9CRxs. We inspected every article manually and confirmed whether we agree with the Allsides rating. Still, since the Allsides rating is rather related to a news outlet than a single news article, it might not represent an exact and complete article bias index. We address the possibilities of extending and improving our article set in Section 5.

### F. Measures

*Perception of Media Bias in articles.* Participants were shown 48 items about the perception of media bias. The included items covered cause (e.g., "Do you think that the article includes different points of view regarding the topic in the article?"), direction (e.g., "This article is… liberal/conservative"), existence (e.g., "This article is biased."), influence (e.g., "How much do you think the news article would influence your view of the issue?") and strength of the bias (e.g., "How biased is the article?"). We measured all answers on VAS with a verbal left and a right anchor.

*Attention check.* To ensure that participants were paying attention, we mixed the item "I read the article" in with the questions on article bias. We anchored the VAS from strongly disagree to agree strongly.

*Perception of General Media Bias.* Six items measured perception of general media bias on a VAS (–10 = strongly disagree to 10 = strongly agree). The different statements about media in general covered the aspects usually referred to in previous research. As a personal variable, we will analyze the results for the perception of general media bias in a different setting and did not consider it for the scale construction described here.

### G. Exploratory factor analysis

To empirically reduce the 48 questions even more and derive a final set of questions that is useable in a single study, we used an exploratory factor analysis (EFA) [37]. An EFA is a statistical technique to reduce data to a smaller set of summary variables and explore and uncover response patterns in survey items. It identifies latent constructs (factors) that define the interrelationship among items by accounting for common variance [37]. An in computer science more widely known special case of an EFA is the Principal Component Analysis (PCA), which uses a linear combination of a set of variables to create one or more index variables. We, however, use the EFA.

The agreement between the survey participants within a EFA can be described in different ways. One of them, which we use in our study, is the Intraclass Correlation (ICC). The ICC is a descriptive statistic that describes how strongly units in the same group resemble each other and can be interpreted as the fraction of variance shared by all raters [38].

While our factor analysis results will allow us to reduce the number of questions reliably, the sample size is not large enough to perform cross-validation. We will therefore run a second validation study in the future, which we address in Section 5.

## IV. Analysis & Results

All articles were rated between one and five times. On average, each article was rated M = 3.49 times (SD = .76). For the factor analysis, we averaged ratings across participants to obtain a mean article rating per item. We computed the agreement between raters per item as the previously described intraclass correlation (ICC) via REML estimates of random intercept models (Table 1) [39].

### A. Factor analysis

The factor analysis used maximum likelihood estimators and oblique promax rotations ($\kappa = 4$). Both KMO (.919) and Bartlett-test ($\chi^2(1128)=9346.38$, $p < .001$) indicated that the selected items were suitable for factor analysis. For determining the number of factors, we used the Velicer's MAP criterium [40], which yielded 6 factors, which could also be viewed as confirmed by the scree-plot (Figure 2). Kaiser criterium yielded 7 and parallel test 5 factors.

TABLE I. Factor analysis with agreement between raters per item as ICC with suppressed loadings below .3. The underlined items were kept for the final question set.

| Items | Rotated factor loadings | | | | | | ICC | Mean (SD) |
|---|---|---|---|---|---|---|---|---|
| | Factuality | Influence | Topic Affirmation | Negative Emotions | Bias | Political Ideology | | |
| <u>In/accurate</u> | <u>0.999</u> | | | | | | 12.64% | 3.25(3.05) |

| | | | | | | | |
|---|---|---|---|---|---|---|---|
| Not/ representing reality | 0.996 | | | | | 14.70% | 3.74(3.32) |
| False/ factual | 0.988 | | | | | 13.24% | 3.83(3.11) |
| Not/ believable | 0.983 | | | | | 10.66% | 4.72(3.10) |
| Un/ trustworthy | 0.978 | | | | | 12.52% | 2.85(3.35) |
| To what extent do you view the article as a credible source of information on the issue? | 0.904 | | | | | 10.26% | 1.97(3.31) |
| Not/ telling the whole story | 0.902 | | | | | 10.69% | 0.98(3.59) |
| Un/informative | 0.882 | | | | | 9.25% | 4.17(3.04) |
| How much do you see the author as trying to communicate information truthfully? | 0.869 | | | | | 13.21% | 3.49(3.26) |
| Un/ethical | 0.853 | | | | | 12.74% | 3.16(3.27) |
| Un/fair | 0.825 | | | | | 13.83% | 2.93(3.49) |
| Im/moral | 0.793 | | | | | 7.48% | 2.78(3.03) |
| Im/balanced | 0.757 | | | | | 18.92% | 1.62(3.66) |
| bad/ good | 0.755 | | | | | 13.42% | 2.30(3.29) |
| Likelihood to read news of the same author | 0.743 | | | | | 4.47% | 0.89(3.39) |
| How informed did you feel as a result of having read the article? | 0.694 | | | | | 8.42% | 1.74(3.27) |
| I would recommend this article to others. | 0.588 | 0.334 | | | | 11.40% | 0.40(3.85) |
| Biased/ unbiased | 0.565 | | | | -0.443 | 11.37% | 0.08(3.72) |
| Do you think the author of this article is trying to change your mind? | -0.528 | 0.34 | | | 0.306 | 10.73% | -1.20(3.77) |
| This article makes me feel sympathetic. | 0.424 | | | | | 23.61% | -0.79(3.35) |
| Do you think that the article includes different points of view regarding the topic in the article? | 0.339 | | | | | 11.91% | -1.45(3.47) |
| How much do you think the news article would influence another person's view of the issue? | | 0.897 | | | | 2.37% | 0.34(2.81) |
| How much influence will the news article you just read have on the attitudes or opinions of people who read this article? | | 0.861 | | | | 4.32% | 0.06(2.84) |
| % changed feelings | | 0.685 | | | | 1.92% | -2.68(2.81) |
| Do you think the article would affect another person's voting behavior? | | 0.684 | | | | 10.70% | -1.11(3.49) |
| This article will impact another person's attitude. | | 0.673 | | | | 0.00% | 1.43(2.69) |
| How much do you think the news article would influence your view of the issue? | 0.304 | 0.643 | | | | 7.53% | -2.42(3.61) |
| This article will impact my attitude. | | 0.489 | | | | 4.80% | -2.33(3.43) |
| Do you think the article would affect your voting behavior? | | 0.449 | | | | 2.35% | -4.88(3.42) |
| The author strongly opposes/supports x | | | 0.952 | | | 48.37% | 0.39(4.62) |
| strongly biased against/towards x | | | 0.908 | | | 32.59% | 0.38(3.54) |
| un/favorable towards x | | | 0.889 | | | 35.41% | 0.44(4.19) |
| Very negative/positive view of x | | | 0.659 | | | 29.67% | -1.16(3.61) |
| Too easy/tough on x | | | -0.438 | | 0.331 | 20.29% | -0.52(2.84) |
| This article makes me feel angry. | | | | 0.997 | | 16.44% | -1.34(4.13) |
| This article makes me feel disgusted. | | | | 0.882 | | 22.12% | -1.48(4.42) |
| This article makes me feel resentful. | | | | 0.758 | | 14.25% | -2.39(3.93) |
| This article makes me feel anxious. | | | | 0.553 | | 14.12% | 2.04(4.07) |
| Reduces/increases political involvement | | | | 0.375 | | 7.13% | 1.03(2.13) |
| To what extent do you perceive that the author's reporting on the issue reflects a bias? | -0.441 | | | | 0.634 | 17.64% | -0.62(3.92) |
| How biased is the article? | -0.475 | | | | 0.634 | 19.93% | -0.58(4.04) |
| Not/ favoring a side | -0.439 | | | | 0.615 | 16.41% | 1.50(3.89) |
| Do you think a reader would perceive the arguments presented in this article to be stronger on one side of the issue than the other? | | | | | 0.594 | 14.23% | 1.49(3.36) |
| Strictly neutral/biased | -0.490 | | | | 0.577 | 19.96% | 0.25(4.24) |
| This article is biased. | -0.507 | | | | 0.540 | 21.52% | -0.66(4.22) |

| | | | |
|---|---|---|---|
| This article makes me feel amused. | | 14.56% | -5.21(3.35) |
| Where would you place the personal view of the author responsible for this article? | 0.965 | 30.15% | -0.98(3.38) |
| liberal/conservative | 0.859 | 29.39% | 0.26(3.40) |

TABLE II. RELIABILITIES OF THE FINAL SCALES BUILT FOR EACH FACTOR AND THEIR CORRELATIONS

| | Factuality | Influence | Topic Affirmation | Negative Emotions | Bias | Political Ideology | Allsides | Ideology$^2$ | Allsides$^2$ |
|---|---|---|---|---|---|---|---|---|---|
| Factuality | .966 | .155* | .050 | -.143* | -.717** | -.273** | -.070 | -.493** | -.328** |
| Influence | | .864 | -.014 | .235** | .098 | -.052 | -.035 | -.064 | -.061 |
| Topic Affirmation | | | .933 | -.221** | -.015 | .169* | .070 | .022 | -.081 |
| Negative Emotions | | | | .887 | .272** | -.054 | -.189** | .203** | .008 |
| Bias | | | | | .953 | .191** | .070 | .520** | .365** |
| Political Ideology | | | | | | .929 | .445** | .016 | .314** |
| Allsides | | | | | | | | .084 | .502** |
| Ideology$^2$ | | | | | | | | | .402** |

As shown in Table 1, the first factor has high loadings on items regarding the factuality of information, the second on perceived influence, the third on the agreement to the topic, the fourth on negative emotion, the fifth on the perceived bias, and the sixth on two items on the political affiliation of the text. Both factor factuality and bias show large cross-loadings. Thus, they may be regarded as facets of a single construct, which also shows in the high correlations between the two scales derived from factors' indicators (Table 2).

The separate interpretation of the factors bias and factuality is motivated by larger differences in the inter-rater agreement for items in the two factors: While factuality seems to have a clearer interpretation of loadings, the bias factor includes items that have considerably larger ICCs. The inter-rater agreement is very low for the factor influence items, indicating that an article's perceived influence is probably widely dependent on a reader-article interaction and therefore not a clear characteristic of an article. It seems that raters agree most on an article's political ideology and whether it affirms a particular topic.

Our basic descriptive analysis of the questions on perception of media bias in articles showed that most mean values were close to the middle of the scale representing medium item difficulty. Four items showed more extreme values. The articles were generally rated as believable and informative. The articles were rated as not amusing or influential on voting behavior. The limited influence on voting behavior might be related to the study being conducted during the US presidential elections. Nevertheless, the items on influence generally showed less dispersion, raising doubts about their suitability to assess a construct on the article level.

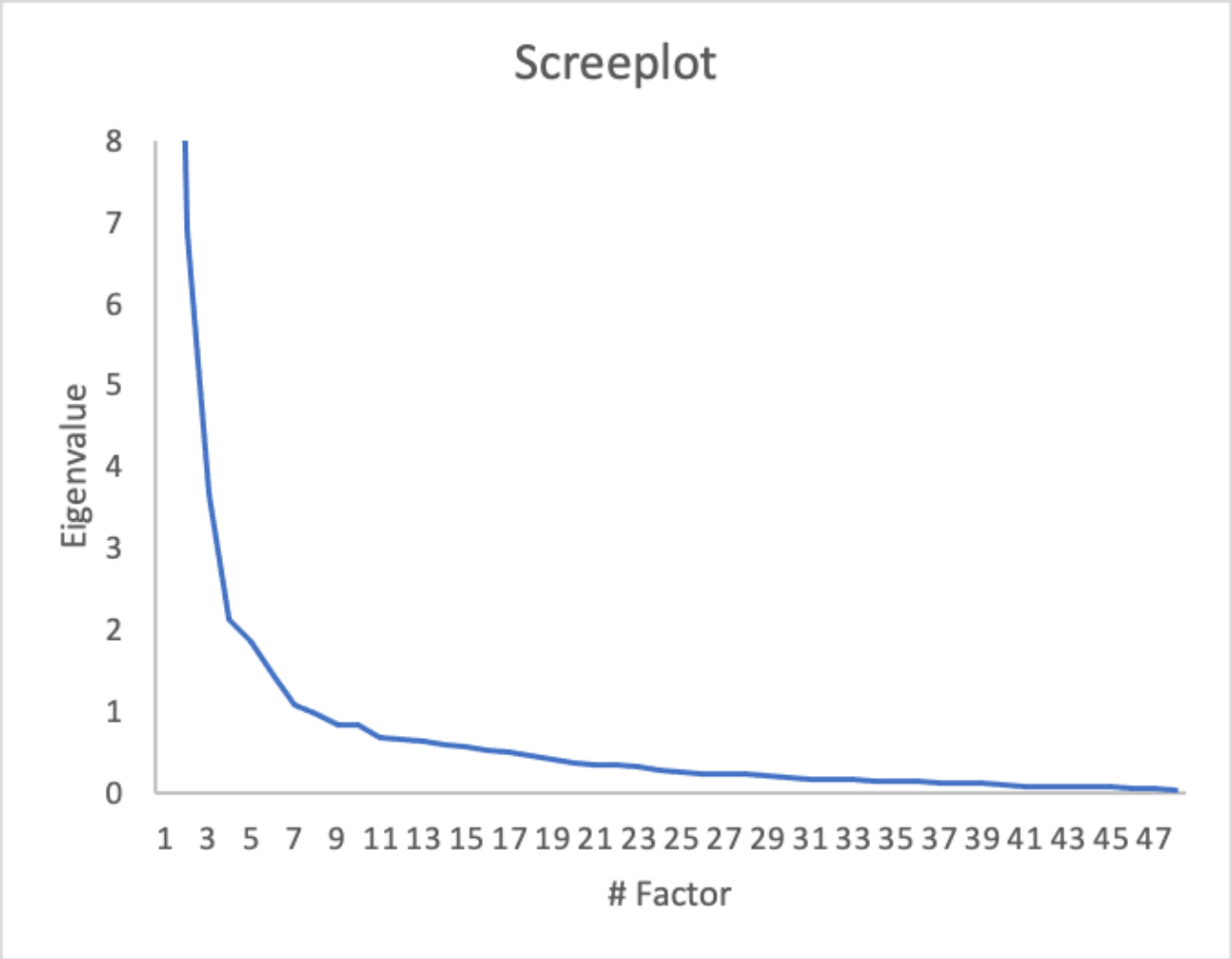

Fig. 2. Screeplot showing the number of factors

To simplify measurement, we decided to use the mean of the items with factor loadings above .7 as scales for each factor. As the scale for factuality would entail too many items, we decided to use a stricter cut-off of .95 for this factor. Likewise, we had to introduce a more liberal cut-off for the factor bias with .5, as all loadings were considerably lower. Dependent on the different cut-off values, we selected the questions for our final scale and question set. We underlined them in Table 1. Overall, we used five items as indicators for factuality, two for influence, three for topic affirmation, three for negative emotionality, six for bias, and two for political ideology. The respective reliabilities of the final preliminary scales (Cronbach's αs) are presented in the diagonals of Table 2 and were acceptable (.89-.97).

## B. *Validation*

Besides the rather large correlation between bias and factuality, as seen in Table 2, all scales are rather independent of each other with small to medium-sized correlations. Comparing the factors to the ratings from Allsides, we see a clear picture that the Allsides rating of ideology mainly correlates with the scale of political ideology. As both the Allsides rating and the scale of political ideology were coded to have lower values for left articles, we also centered and squared both scales to obtain a measure of political extremity with lowest values for politically neutral articles and highest values for both very left and right articles. These computed measures of political extremity yielded medium correlations with both bias and factuality scales. In sum, there is the first indication of the validity of our article-specific scales of bias, factuality, and political ideology.

## V. Discussion

Our work's main goal was to develop a reliable scale to evaluate media bias in news articles further and improve existing data sets and questionnaires. Our search for items resulted in a list of 419 items from 74 different papers on media bias perception. To the best of our knowledge, such a collection of items is the most sophisticated on the issue to date and even the first of its kind. The items chosen for the questionnaire evaluated in this study covered the different areas and nuances within the different items while limiting the number to a testable amount of 48 questions.

The exploratory factor analysis resulted in a structure with six interpretable factors: factuality, influence, topic affirmation, negative emotion, bias, and political ideology. While most of them were independent of each other, marked by low correlations, the factors bias and factuality were highly interrelated. They thus may also be regarded as subfacets of one construct in future cross-validation attempts.

For a benchmark for automatic classifiers, the factors bias and political ideology (potentially squared) seem particularly useful as they mostly tap into our concept of bias. Using the political ideology factor may be most efficient, as raters seem to agree more on this dimension than the bias factor. Both the factuality factor and the negative emotions factor could further contextualize the ratings of the articles.

On the other hand, the scale of influence was rather rater dependent and thus seems less suitable for generalizable ratings on the article level. Similarly, the topic affirmation factor is a bit context-dependent, as one has to decide how to define each article's topic separately. However, an argument in favor of using this factor for bias assessment may be its very high inter-rater agreement.

The medium correlations with political ideology with the external Allsides rating may be seen as a validation for the rated items. It also suits our concept of perceived media bias that the extremity of the political ideology was considerably correlated with both the bias and the factuality factors. Please note that the correlations with the Allsides ratings are also deflated, as Allsides provides a rating only for the media outlet, but not the actual article. While we believe that the ratings and our manual article inspection offer reasonable ground to measure rater agreement, we will further improve our data set in the future.

When interpreting the results of our study, a few limitations have to be taken into consideration. At first, the item collection, categorization, and reduction process was only performed by one person, potentially leading to an implicit bias in item selection. To counteract the problem, the entire process was meticulously documented and is transparently visualized on http://bias-question-tree.gipplab.org/. Furthermore, the factors identified were a result of an exploratory factor analysis. In the future, the factor structure should thus be validated with a different set of articles potentially using additional external validation criteria. Likewise, the suggested six scales with 21 items derived from the factor analysis are somewhat preliminary and subject to further testing and validation. Finally, our questionnaire is designed to detect media bias in articles. In today's world, many people get their news from video and audio clips. While we believe it is reasonable to assume that many questions from this questionnaire could also be used in a study on different formats, some aspects could still be medium-specific.

Despite the limitations, our study is an important step towards an improved understanding of media bias and its perception. Our results suggest that the selected items are a good foundation for creating a final questionnaire on media bias, which will be our main focus for future work. Different factors influencing media bias perception can be studied more easily by asking participants additional and reliable questions. This study can help researchers from different fields by providing them with a good tool to measure media bias. For example, our questionnaire could be used to compare and validate an automated bias classifier with human assessments (and vice versa) on a more reliable level than before. Since perception is subjective [14], the comparison of results of this questionnaire to automated ratings could be a valuable insight into the dimensions of this subjectivity. Overall, we believe that standardized and evaluated questions will be an important step for researching media bias and related areas of interest.

## VI. Conclusion

The perception of media bias is an important research area because of its strong relation to collective decision-making and communication processes. Our study summarizes and evaluates prior research to create a scale and standard questionnaire on media bias for future applications. The results show that the perception of media bias is a multi-faceted construct, influenced by factors like political orientation. The basic analysis of the items we selected and their comparison to the AllSides bias-ratings suggest that the items chosen are adequate for measuring media bias perception with an acceptable amount of agreement among raters. The scales constructed and analyzed in this study thus provide a basis to create a standard tool for measuring media bias.